\pdfoutput=1

\documentclass[graybox, envcountchap]{svmult}

\usepackage{mathptmx}       
\usepackage{helvet}         
\usepackage{courier}        
\usepackage{type1cm}        
\usepackage{makeidx}         
\usepackage{graphicx}        
\usepackage{multicol}        
\usepackage[bottom]{footmisc}
\usepackage{adjustbox}
\usepackage{tabulary}
\usepackage{dirtytalk}

\makeindex             
\usepackage{algorithm} 
\usepackage{algpseudocode} 
\usepackage{multirow}

\begin{document}

\title*{Enhancing Self-Disclosure In Neural Dialog
Models By Candidate Re-ranking}
\author{Mayank Soni, Benjamin R. Cowan and Vincent Wade}
\institute{Mayank Soni \at ADAPT Centre, Trinity College Dublin, \email{sonim@tcd.ie}
\and Benjamin R. Cowan \at ADAPT Centre, University College Dublin \email{benjamin.cowan@ucd.ie}
\and Vincent Wade \at ADAPT Center, Trinity College Dublin, \email{vincent.wade@adaptcentre.ie}}
%
%
\maketitle



\abstract{Neural language modelling has progressed the state-of-the-art in different downstream Natural Language Processing (NLP) tasks. One such area is of open-domain dialog modelling, neural dialog models based on GPT-2 such as DialoGPT have shown promising performance in single-turn conversation. However, such (neural) dialog models have been criticised for generating responses which although may have relevance to the previous human response, tend to quickly dissipate human interest and descend into trivial conversation. One reason for such performance is the lack of explicit conversation strategy being employed in human-machine conversation. 
Humans employ a range of conversation strategies while engaging in a conversation, one such key social strategies is \textit{Self-disclosure} (SD). A phenomenon of revealing information about one-self to others. 
In this work, Self-disclosure enhancement architecture (SDEA) is introduced utilizing Self-disclosure Topic Model (SDTM) during inference stage of a neural dialog model to re-rank response candidates to enhance self-disclosure in single-turn responses from from the model.}

\section{Introduction}
\label{sec:1}
Neural Language models based on neural language pre-training such as GPT \cite{radford2019language}, GPT-2 \cite{radford2018improving} has advanced the state-of-the-art in various NLP Tasks. 
Open-domain dialogue models such as DialoGPT \cite{zhang2020dialogpt}, Blender-bot \cite{roller2020recipes} and Meena \cite{adiwardana2020towards} have shown promising performance in single-turn conversation. However, when focused on social talk, such (neural) deep learning solutions have been strongly criticised for generating conversational utterances which although may have relevance to the previous human utterance, tend to quickly dissipate human interest and descend into trivial conversation and disengage with the human user \cite{huang2020challenges}, \cite{zhang2018personalizing}, \cite{li2015diversity}.  Some of the problems in their usage result in generating bland and repetitive responses like \say{Thank you}, \say{ok}, \say{I don’t know} \cite{huang2020challenges}, \cite{gao2018neural}, \cite{vinyals2015neural}.  

One of the reasons for such responses is the lack of explicit conversation strategy employed by a neural dialog system. Hence, in this paper we seek to enhance existing neural conversational agents based on novel model adaptation which enables the neural dialog model to activate \textit{self-disclosure}. Self-disclosure has been well researched as higher order conversational skill in psychology literature where human centric evaluation of such strategies has proven to enhance the relationship between interlocutors and positively affect their engagement in the conversation \cite{zhao-etal-2016-automatic}, \cite{jain2018user}, \cite{ravichander-black-2018-empiricalRavichander}. However, to the authors knowledge, attempting to explicitly empower a neural dialog model to utilise self-disclosure with a corpus-neutral approach has not been attempted before. In this paper we focus on integrating Self-disclosure and evaluate the degree to which it is seen enhanced in the responses by implementing a novel architecture. Below, we briefly describe self-disclosure level, Self-disclosure Topic Model (SDTM), Self-disclosure enhancement architecture (SDEA), experiment and results.


\section{Self-Disclosure}
Humans employ a range of conversation strategies to fulfil multiple goals in a conversation \cite{tracy1990multiple}. Conversation strategies are discourse units which could span multiple utterance turns and are typically larger than speech acts \cite{zhao-etal-2016-automatic}. Such strategies contribute to building and maintaining human relationships during the dialogue. Humans employ a range of different strategies to build rapport and increase interpersonal cohesiveness \cite{zhao-etal-2016-automatic}. It is argued that over time humans behave in ways to fulfill mutual behaviour expectations \cite{spencer2008culturally}. One of the most important conversation strategies is \textit{Self-disclosure}. Researchers \cite{altman1973} have expounded on the idea of social penetration, Social Penetration Theory (SPT) proposes that communication between two people moves from shallow to deeper levels as the relationship progresses. SPT primarily proposes the idea that relationships progress through self-disclosure, a phenomenon of revealing information about one-self to others. This information about oneself could consist of one’s thoughts, aspirations, past events and future plans. Self-disclosure is a key social strategy identified by information being revealed about oneself. This helps in creating rapport and a feeling of mutual trust among the participants engaging in dialogue. People disclose information about themselves to improve and maintain relationships and form deeper connections. Employing appropriate self-disclosure can lead to a feeling of mutual trust, friendliness and overall satisfaction in a conversation \cite{joinson2007self}. Researchers showed that self-disclosure is a key strategy in building rapport in a peer-to-peer tutoring experiment \cite{zhao-etal-2016-automatic}. It has also been shown that self-disclosure in a human-machine conversation can lead to higher satisfaction in conversation \cite{jain2018user},\cite{ravichander-black-2018-empiricalRavichander}. Motivated by research in psycho-linguistics, socio-linguistics and SPT, self-disclosure enhancement architecture (SDEA) in neural dialog systems is implemented and evaluated.


\subsection{Self-disclosure ($SD$) level}
\label{sec:2}
 
Self-disclosure can be divided into multiple levels. We follow the three level self-disclosure recognition as highlighted in \cite{bak-etal-2014-self}. The study highlights three levels of self-disclosure from social science and psychology literature: $G$(general) for no-disclosure, $M$ (medium disclosure) and $H$ (high disclosure) \cite{vondracek1971manipulation}, \cite{barak2007degree}. These three levels are organized in progressing order of sensitive information being revealed by an agent. 

($G$) levels of self-disclosure includes no self-disclosure. Responses that are about a third-person, event or thing are labelled as general disclosure.($M$) level of self-disclosure comprise of information about oneself. Examples are statement that increase information about a user such as birthday, events etc. Personal pronouns such as \lq{My}\rq{}, \lq{I}\rq{} are identifiers of medium level disclosure in an utterance. Medium disclosure contains information that is non-sensitive in nature. $H$ levels of self-disclosure contains personal and sensitive information. Mentioning concerns and insecurities is a cue to identify high levels of self-disclosure. Responses such as \textit{\lq{I am overweight and trying to loose some weight}\rq{}} is an example of high self-disclosure. We refer to \cite{bak-etal-2014-self} for list of the keywords that help identifying $H$ levels of self-disclosure.


\subsection{Self-disclosure recognition model}
 The first step towards re-ranking response candidates, described in section \ref{section4}, is to be able to recognise self-disclosure levels computationally. There are various researches which have implemented a supervised classifier \cite{ravichander-black-2018-empiricalRavichander}, \cite{diyi_sd}, however these require disclosure annotated dataset to train a classifier. Hence, we utilize SDTM \cite{bak-etal-2014-self} as it is a semi-supervised model to recognise levels of self-disclosure as discussed in section \ref{sec:2}. This was developed for recognising levels of self-disclosure in longitudinal Twitter conversations. The model recognises three levels of self-disclosure as mentioned in section \ref{sec:2} and relies on seed-keywords and n-grams to recognise the level of self-disclosure. The model classifies a given sentence into $G$ vs $M/H$ of disclosure. $G$ means no disclosure and $M/H$ are medium and high level disclosures. $M$ level (non-sensitive) disclosures are defined by referring to personal pronouns and sharing information about oneself. While high level self-disclosure is classified by revealing secret or vulnerable information about oneself. The classification of degree of self-disclosure into High, medium and general is based on \cite{vondracek1971manipulation} and \cite{barak2007degree}. $FP+SE1$ is utilized to recognise self-disclosure levels from SDTM \cite{bak-etal-2014-self}.

\section{Self-Disclosure Enhancement By Candidate Re-ranking}
\label{section4}

Neural Language models based on neural language pre-training such as GPT \cite{radford2019language}, GPT-2 \cite{radford2018improving} has advanced the state-of-the-art in various NLP Tasks. Pre-training has advanced the development of neural models for open-domain conversation generation. DialoGPT \cite{zhang2020dialogpt} was released as a pre-trained dialog response generation model based on GPT-2. At the core of DialoGPT \cite{zhang2020dialogpt} is language modelling, a task of estimating unsupervised language distribution from a set of examples ($t_1, t_2, t_3,...,t_n$). Since open-domain dialog follows a natural sequence, turns in dialog can be modelled as product of conditional probabilities. If source sentence $S$ consists of a series of tokens ($t_1, t_2, t_3,...,t_m$). Then, the response sentence $R$ can be framed as continuation of source tokens ($t_(m+1),...,t_n$) 

\begin{equation}
\label{equation_1}
    p(R|S) = \prod_{m+1}^n P(t_n|t_1,...,t_{n-1})
\end{equation}

DialoGPT is employed as the base neural dialog model for experimentation. Typically neural dialog models (such as DialoGPT) consist of three components namely Training Corpus, Neural Architecture and Inference Strategy. This paper focuses on adaptation in the Inference Stage of a neural dialog system. Neural dialog models generate responses by following a probability or sampling based inference strategy. Inference strategies in open-domain dialog systems are based on probabilistic sampling (e.g. nucleus sampling \cite{holtzman2020curious}, top-k \cite{fan-etal-2018-hierarchical}, beam search, greedy sampling) from a fixed vocabulary. Rather than optimizing for semantic coherence, this work investigates if adaptation of the inference mechanism should be explored to consider semantically coherent, lower probability responses that are more indicative of a self-disclosure level ($G$, $M$, $H$). To evaluate this hypothesis, Self-Disclosure Topic Model (SDTM) \cite{bak-etal-2014-self} is used to search response candidates for a pre-defined self-disclosure level and the said candidate is rendered as the response. Figure 1 illustrates the architecture of SDEA. The gray unit in Figure 1 provides an overview of SDEA. Response (yellow) and SDEA Response(green) units show handpicked example of response generation from DialoGPT \cite{DialoGPT2-Interact} and DialoGPT enhanced with SDEA.

\begin{algorithm}[htbp!]
    \label{algorithm}
	\caption{Self-Disclosure Enhancement Architecture (SDEA)}
	\begin{algorithmic}[1]
		\For {$t$ in range ($sequence length$) : sample tokens ($t_1, t_2,...,t_n$})
		         \EndFor
		\State Join tokens to form one sequence $S$ 
		\State Split $S$ on $eos -id$ to obtain candidates $C_1, C_2,...,C_m$
		\State Compute $SD$ level of candidates $C_1, C_2,...,C_m$
		\State Render 1st candidate in sequence $S$, $C_x$ with specified $SD$ level 
	\end{algorithmic} 
\end{algorithm}

The decoding algorithm generates tokens defined by the sequence length. For instance if the sequence length is $20$, $20$ tokens will be generated in accordance with the decoding strategy chosen. The tokens also contain end-of-text ($eos-id$) token which signifies the end of sentence. This leads to generation of multiple sentences separated with a end-of-text token within a sequence. These are called response candidates. Theoretically, we argue that given a large distribution of tokens and a large dataset. It should be possible to generate responses from a neural dialog system consisting of a certain SD level. Algorithm 1 elaborates steps involved in generating SD enhanced response.

\begin{figure*}[htbp!]
  \includegraphics[scale=0.45]{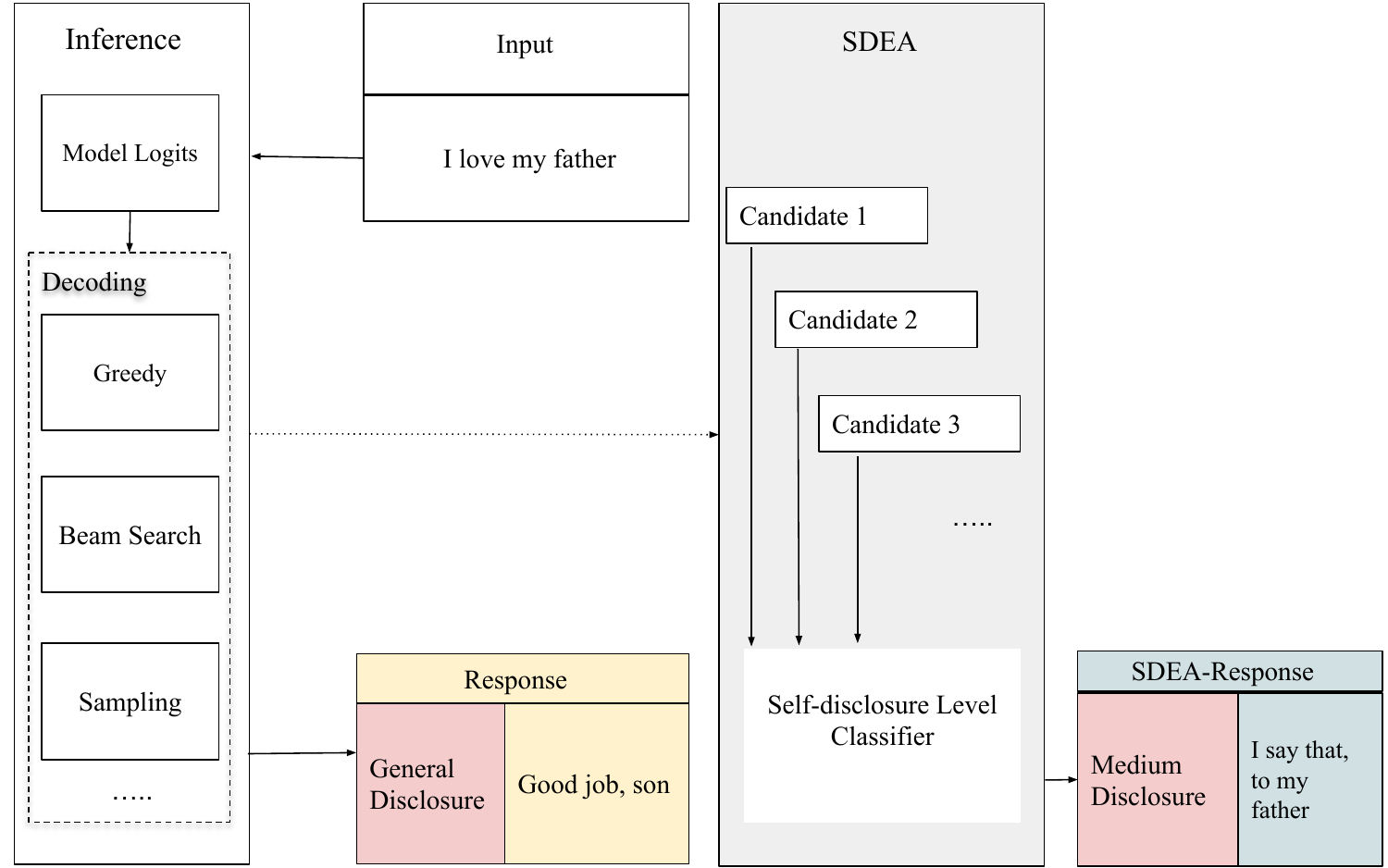}
  
  \label{figure3}
  \centering
    \caption{Overview of SDEA (grey unit). Response in green is Self-disclosure enhanced response }
\end{figure*}


\section{Experiment}
The central objective of the experiment was to evaluate, if the self disclosure levels (as defined in \ref{sec:2}) are higher than that of a vanilla system. Vanilla responses are defined as responses generated by a neural dialog system using a decoding strategy, DialoGPT \cite{DialoGPT2-Interact} in this case. To evaluate the effect of the proposed self-disclosure enhancement in neural dialog model by re-ranking candidates architecture, responses from vanilla and self-disclosure enhancement in neural dialog model by re-ranking candidates architecture are generated.

Medium sized DialoGPT model is employed as the Neural dialog system for the experiment. Changes are made to the decoding script from \cite{DialoGPT2-Interact}. Vanilla responses are generated with Nucleus sampling ($top-p$) value of $0.9$ and a sequence length of $100$. Self-disclosure enhanced responses are generated by incorporating SDEA with aforementioned vanilla response setup. A larger sequence length is used so that enough candidates could be produced to make a selection from. The experiment is performed on two dialog datasets: Dailydialog \cite{li2017dailydialog} and Switchboard \cite{switchboard}. The description of dataset preparation is in section \ref{data_description}. Following the dataset preparation, responses from both (SDEA and Vanilla system), with their SD levels is obtained. Finally, Chi-Square test is conducted on the the distribution of SD levels from both the systems. The contingency table with Chi-square values and $p$ values can be seen in Table under Figure 3. Results are further discussed in section \ref{result}. The decoding script from \cite{DialoGPT2-Interact} is utilized. The history is erased after each response generation so that there is no affect of context on the new turn.

\subsection{Data Configuration}
\label{data_description}
Data is prepared using two dialog datasets viz. Dailydialog \cite{li2017dailydialog} and Switchboard \cite{switchboard}. Since, in the early stages of a conversation, self-disclosure leads to longer conversations \cite{ravichander-black-2018-empiricalRavichander}, and the task is to test the enhancement of self-disclosure in response from a neural dialog model in a single turn. Only the first dialog turn from each conversation in the dataset is selected. This leads to creation of a dataset consisting of only first (or second if the first dialog sentence is noisy) dialog turn. Handpicked examples of such dialog turns from both, Dailydialog \cite{li2017dailydialog} and Switchboard \cite{switchboard} can be seen in table 1.


\begin{table*}
    \label{example_commands}
    \begin{tabulary}{\textwidth}
    {l | l}
    \hline
    \\
    \textbf{Dailydialog} & \textbf{Switchboard}
    \\
    \\\hline \\
    Are things still going badly with your house guest &  
    Um how's it been this week for you  \\ \\
    
    What kind of food do you like & all right Amy how are you doing today  \\ \\ 
    
    Hello is John in & so who's your uh favorite team  \\ \\ 
    
    \hline

    \end{tabulary}

\caption{Handpicked first prompts examples from Dailydialog and Switchboard}

\end{table*}

\subsection{Result \& Error Analysis}
\label{result}
The results in table under Figure 3 show a clear difference in distribution of responses between general and medium disclosure in both datasets, indicating that SD levels are higher from SDEA. In the DailyDialog dataset, 39.20\% of responses from vanilla system have \textit{medium} level self-disclosure, whereas 97.60\% responses from the self-disclosure enhancement system have \textit{medium} level self-disclosure. Similarly, out of 1277 prompts in the switchboard dataset, 33.90\% had medium disclosure from the vanilla system, whereas 95.20\% responses have \textit{medium} disclosure levels from SDEA. Also, SDEA was unable to find \textit{medium} disclosure responses for 4.80\% switchboard prompts and 2.40\% prompts because no candidate with a medium disclosure was found within a sequence length of 100. Pearson's Chi-Square test is then performed to confirm that the SD level distributions from the SDEA and Vanilla systems are statistically significant. The $p$ values for both datasets reveal that the SD levels are distributed differently, and  $DialoGPT +SDEA$ responses clearly lean towards generating responses with medium disclosure. Results can be seen in Table 2 and Table 3.




\begin{table*}[htbp!]
\caption{Automatic dialog evaluation metrics scores on filtered DailyDialog dataset}       
\newcolumntype{M}[1]{>{\arraybackslash}m{#1}}
\resizebox{\textwidth}{!}{%
    \centering
    \begin{tabular}{
    M{1.6cm}                
    M{1.5cm}
    M{1.3cm}               
    M{1cm}
    M{1cm}                
    M{1cm}
    M{1cm}               
    M{1.5cm}
    M{0.75cm}                
    M{0.75cm}
    M{1cm}}

\hline 
{Method}  & \multicolumn{2}{c}{SD Level} & \multicolumn{2}{c}{NIST} & \multicolumn{2}{c}{BLEU}& ENTROPY & \multicolumn{2}{c}{DIST}&{Avg. Length} \\

 & General&Medium&N-2&N-4&B-2&B-4&E-4&D1&D2 \\
 \hline
 
DIALOGPT\\(Nucleus Sampling, p=0.9)&60.80\%&39.20\%&\textbf{0.37}&\textbf{0.37}&\textbf{0.016}&\textbf{0.003}&8.83&\textbf{0.23}&\textbf{0.75}& 09.85

 \\\hline
 
 DIALOGPT \\(Nucleus Sampling, p=0.9) \textbf{+SDEA}&02.40\%&\textbf{97.60}\%&0.33&0.33&0.009&0.001&\textbf{8.98}&0.17&0.67&\textbf{11.03}

\\\hline
 
\end {tabular}}
\end{table*}

\begin{table*}[htbp!]
\caption{Automatic dialog evaluation metrics scores on filtered Switchboard dataset}
\label{tab:sb} 
\newcolumntype{M}[1]{>{\arraybackslash}m{#1}}
\resizebox{\textwidth}{!}{%
    \centering
    \begin{tabular}{
    M{1.6cm}                
    M{1.5cm}
    M{1.3cm}
    M{1cm}
    M{1cm}            
    M{1cm}
    M{1cm}  
    M{1.5cm}
    M{0.75cm}       
    M{0.75cm}
    M{1cm}}

\hline
{Method}  & \multicolumn{2}{c}{SD Level} & \multicolumn{2}{c}{NIST} & \multicolumn{2}{c}{BLEU}& ENTROPY & \multicolumn{2}{c}{DIST}&{Avg. Length} \\

 & General&Medium&N-2&N-4&B-2&B-4&E-4&D1&D2 \\
 \hline 
 
DIALOGPT \\(Nucleus Sampling, p=0.9)&66.10\%&33.90\%&0.27&0.27&0.01&0.0005&9.10&\textbf{0.22}&\textbf{0.73}&10.17

 \\ \hline
 
 DIALOGPT \\(Nucleus Sampling, p=0.9) \textbf{+SDEA} &4.80\%&\textbf{95.20}\%&\textbf{0.34}&\textbf{0.34}&0.01&0.0005&\textbf{9.22}&0.17&0.67&\textbf{10.94}

\\\hline 
 
\end {tabular}}

\end{table*}

We further evaluate responses from DialogGPT and DialogGPT +SDEA on various automated dialog metrics such as BLEU \cite{papineni2002bleu}, NIST \cite{doddington2002automatic}, METEOR \cite{lavie2007meteor}, Entropy \cite{zhang2018generating}, Dist-n \cite{li2015diversity}. The primary reason for this evaluation is to be caution against irrelevancy when enhancing self-disclosure. It is observed that the difference between the aforementioned systems is minimal. For DailyDialog dataset, Vanilla system has better NIST, BLEU and DIST scores and Vanilla System +SDEA has better Entropy and Avg.Length. Similarly, for Switchboard dataset, Vanilla system has better DIST scores and Vanilla System +SDEA has better NIST, Entropy and Avg. Length. Thus, it can be inferred that enhancing self-disclosure does not lead to irrelevant response generation. Future work will include evaluation in Multi-reference setting \cite{gupta2019investigating} using reddit multi-reference dataset \cite{zhang2020dialogpt}.

\section{Discussion, Limitation and Future Work}

This study highlights that the current decoding strategies do not, yet, take into account relationship building with the user by employing any method such as using a conversation strategy. One of the limitation of the proposed method is that the processing is post-generation. Hence, we are limited by the candidates generated by a Language Model Decoding Strategy. If the candidates within a given sequence length do not have a specified level of disclosure then there would be no disclosure enhanced response generated. Hence, future work will include changes to training stage to generate enhanced disclosure responses directly. Human evaluation of the improved self-disclosure and disclosure reciprocity will be part of future research.


\section{Conclusion}
SDEA was presented as an architecture to enhance self-disclosure in neural dialog systems by response candidate re-ranking. The approach under consideration was corpus-neutral, meaning that no changes to the corpora were made, and the only change was made during the inference stage. 
This is helpful since the technique may be employed with a variety of decoding-based neural dialog models. A step is  taken in the direction of making dialog systems conversation strategy aware.

\section{Acknowledgement}
This work was conducted with the financial support of the Science Foundation Ireland Centre for Research Training in Digitally-Enhanced Reality (D-REAL) under Grant No. 18/CRT/6224. We would like to thank anonymous reviewers from IWSDS 2021 for their valuable comments.


\begin{thebibliography}{99.}%
%


\bibitem{joinson2007self} 
Joinson, A. N., Paine, C. B. (2007). Self-disclosure, privacy and the Internet. The Oxford handbook of Internet psychology, 2374252.







\bibitem{altman1973}
Altman, I.,  Taylor, D. (1973). DA (1973): Social penetration: The development of interpersonal relationships. Holt, New York.















\bibitem{bak-etal-2014-self}
Bak, A. (2014). Self-disclosure topic model for classifying and analyzing Twitter conversations. In Proceedings of the 2014 Conference on Empirical Methods in Natural Language Processing (EMNLP) (pp. 1986–1996). Association for Computational Linguistics.



\bibitem{ravichander-black-2018-empiricalRavichander}, A. (2018). An Empirical Study of Self-Disclosure in Spoken Dialogue Systems. In Proceedings of the 19th Annual SIGdial Meeting on Discourse and Dialogue (pp. 253–263). Association for Computational Linguistics.



\bibitem{zhao-etal-2016-automatic}
Ravichander, A. (2018). An Empirical Study of Self-Disclosure in Spoken Dialogue Systems. In Proceedings of the 19th Annual SIGdial Meeting on Discourse and Dialogue (pp. 253–263). Association for Computational Linguistics.


\bibitem{vondracek1971manipulation}
Vondracek, S.,  Vondracek, F. (1971). The manipulation and measurement of self-disclosure in preadolescents. Merrill-Palmer Quarterly of Behavior and Development, 17(1), 51–58.


\bibitem{barak2007degree}
Barak, A., Gluck-Ofri, O. (2007). Degree and reciprocity of self-disclosure in online forums. CyberPsychology and Behavior, 10(3), 407–417.










\bibitem{diyi_sd}
Yang, D., Yao, Z., Kraut, R. (2017, May). Self-disclosure and channel difference in online health support groups. In Proceedings of the International AAAI Conference on Web and Social Media (Vol. 11, No. 1).

\bibitem{zhang2020dialogpt}
Zhang, Y., Sun, S., Galley, M., Chen, Y. C., Brockett, C., Gao, X., ...  Dolan, B. (2019). Dialogpt: Large-scale generative pre-training for conversational response generation. arXiv preprint arXiv:1911.00536.
      


\bibitem{roller2020recipes}
Stephen Roller, Emily Dinan, Naman Goyal, Da Ju, Mary Williamson, Yinhan Liu, Jing Xu, Myle Ott, Kurt Shuster, Eric M. Smith, Y-Lan Boureau,  Jason Weston. (2020). Recipes for building an open-domain chatbot.







\bibitem{fan-etal-2018-hierarchical}
Fan, Y. (2018). Hierarchical Neural Story Generation. In Proceedings of the 56th Annual Meeting of the Association for Computational Linguistics (Volume 1: Long Papers) (pp. 889–898). Association for Computational Linguistics.


\bibitem{vinyals2015neural}
Oriol Vinyals, Quoc Le. (2015). A Neural Conversational Model.






\bibitem{radford2019language}
Radford, A., Wu, J., Child, R., Luan, D., Amodei, D.,  Sutskever, I. (2019). Language models are unsupervised multitask learners. OpenAI blog, 1(8), 9.


\bibitem{radford2018improving}
Radford, A., Narasimhan, K., Salimans, T.,  Sutskever, I. (2018). Improving language understanding by generative pre-training.



\bibitem{DialoGPT2-Interact}
andreamad8. (n.d.). andreamad8/DialoGPT2-Interact. GitHub. https://github.com/andreamad8/DialoGPT2-Interact. 


\bibitem{holtzman2020curious}
Ari Holtzman, Jan Buys, Li Du, Maxwell Forbes,  Yejin Choi. (2020). The Curious Case of Neural Text Degeneration.






\bibitem{spencer2008culturally}
Spencer-Oatey, H. (2008). Culturally Speaking Second Edition: Culture, Communication and Politeness Theory. Bloomsbury Publishing.




\bibitem{tracy1990multiple}
Tracy, K.,  Coupland, N. (1990). Multiple goals in discourse: An overview of issues. Journal of Language and Social Psychology, 9(1-2), 1–13.




\bibitem{jain2018user}
Jain, A., Pecune, F., Matsuyama, Y.,  Cassell, J. (2018). A user simulator architecture for socially-aware conversational agents. In Proceedings of the 18th International Conference on Intelligent Virtual Agents (pp. 133–140).

\bibitem{adiwardana2020towards}
Adiwardana, D., Luong, M.T., So, D., Hall, J., Fiedel, N., Thoppilan, R., Yang, Z., Kulshreshtha, A., Nemade, G., Lu, Y.,  others (2020). Towards a human-like open-domain chatbot. arXiv preprint arXiv:2001.09977.


\bibitem{huang2020challenges}
Huang, M., Zhu, X., Gao, J. (2020). Challenges in building intelligent open-domain dialog systems. ACM Transactions on Information Systems (TOIS), 38(3), 1–32.


\bibitem{gao2018neural}
Gao, J., Galley, M., Li, L. (2018). Neural approaches to conversational ai. In The 41st International ACM SIGIR Conference on Research and Development in Information Retrieval (pp. 1371–1374).



















\bibitem{zhang2018personalizing}
Zhang, S., Dinan, E., Urbanek, J., Szlam, A., Kiela, D., Weston, J. (2018). Personalizing dialogue agents: I have a dog, do you have pets too?. arXiv preprint arXiv:1801.07243.






\bibitem{holtzman2019curious}
Holtzman, A., Buys, J., Du, L., Forbes, M., Choi, Y. (2019). The curious case of neural text degeneration. arXiv preprint arXiv:1904.09751.


\bibitem{li2017dailydialog}
Yanran Li, Hui Su, Xiaoyu Shen, Wenjie Li, Ziqiang Cao, Shuzi Niu. (2017). DailyDialog: A Manually Labelled Multi-turn Dialogue Dataset.

\bibitem{switchboard}
Calhoun, S., Carletta, J., Brenier, J. M., Mayo, N., Jurafsky, D., Steedman, M., Beaver, D. (2010). The NXT-format Switchboard Corpus: a rich resource for investigating the syntax, semantics, pragmatics and prosody of dialogue. Language resources and evaluation, 44(4), 387-419.

\bibitem{papineni2002bleu}
Papineni, K., Roukos, S., Ward, T., Zhu, W. J. (2002, July). Bleu: a method for automatic evaluation of machine translation. In Proceedings of the 40th annual meeting of the Association for Computational Linguistics (pp. 311-318).

\bibitem{doddington2002automatic}
Doddington, G. (2002, March). Automatic evaluation of machine translation quality using n-gram co-occurrence statistics. In Proceedings of the second international conference on Human Language Technology Research (pp. 138-145).

\bibitem{lavie2007meteor}
Lavie, A., Agarwal, A. (2007, June). METEOR: An automatic metric for MT evaluation with high levels of correlation with human judgments. In Proceedings of the second workshop on statistical machine translation (pp. 228-231).

\bibitem{zhang2018generating}
Zhang, Y., Galley, M., Gao, J., Gan, Z., Li, X., Brockett, C., Dolan, B. (2018). Generating informative and diverse conversational responses via adversarial information maximization. arXiv preprint arXiv:1809.05972.


\bibitem{li2015diversity}
Li, J., Galley, M., Brockett, C., Gao, J. and Dolan, B., 2015. A diversity-promoting objective function for neural conversation models. arXiv preprint arXiv:1510.03055.


\bibitem{gupta2019investigating}
Gupta, P., Mehri, S., Zhao, T., Pavel, A., Eskenazi, M. and Bigham, J.P., 2019. Investigating evaluation of open-domain dialogue systems with human generated multiple references. arXiv preprint arXiv:1907.10568.
















\end{thebibliography}
\end{document}